\documentclass{bmvc2k}

\usepackage{amsmath}
\usepackage{amssymb}
\usepackage{color, colortbl}
\usepackage{algpseudocode,algorithm,algorithmicx}
\usepackage[normalem]{ulem}
\usepackage{lipsum}

\newcommand\blfootnote[1]{%
  \begingroup
  \renewcommand\thefootnote{}\footnote{#1}%
  \addtocounter{footnote}{-1}%
  \endgroup
}
\useunder{\uline}{\ul}{}
\newcommand*\Let[2]{\State #1 $\gets$ #2}
\title{Motion-Bias-Free Feature-Based SLAM}

\addauthor{Alejandro Fontan}{fontanvillacampa@gmail.com}{1}
\addauthor{Javier Civera}{jcivera@unizar.es}{2}
\addauthor{Michael Milford}{michael.milford@qut.edu.au}{1}

\addinstitution{
 QUT Centre for Robotics\\
 Queensland University of Technology\\
 Brisbane, Australia
}
\addinstitution{
School of Engineering\\
 University of Zaragoza\\
 Zaragoza, Spain
}

\runninghead{Fontan, Civera, Milford}{Motion-Bias-Free Feature-Based SLAM }

\begin{document}

\maketitle

\begin{abstract}

For SLAM to be safely deployed in unstructured real world environments, it must possess several key properties that are not encompassed by conventional benchmarks. In this paper we show that SLAM \emph{commutativity}, that is, consistency in trajectory estimates on forward and reverse traverses of the same route, is a significant issue for the state of the art. Current pipelines show a significant bias between forward and reverse directions of travel, that is in addition inconsistent regarding which direction of travel exhibits better performance. In this paper we propose several contributions to feature-based SLAM pipelines that remedies the motion bias problem. In a comprehensive evaluation across four datasets, we show that our contributions implemented in ORB-SLAM2 substantially reduce the bias between forward and backward motion and additionally improve the aggregated trajectory error. Removing the SLAM motion bias has significant relevance for the wide range of robotics and computer vision applications where performance consistency is important. 

\end{abstract}

%-------------------------------------------------------------------------
\blfootnote{This research was partially supported by funding from ARC Laureate Fellowship FL210100156 to MM and the QUT Centre for Robotics, the Spanish Government under Grant PID2021-127685NB-I00 and TED2021-131150B-I00 and the Aragon Government under Grant DGA\_FSE\-T45\_20R. }

\section{Introduction}
\label{sec:introduction}
\textbf{Simultaneous Localization and Mapping (SLAM)} refers to the concurrent estimation of a mobile robot state and a model of the environment, known as the \textit{map}, from its onboard sensor readings. Visual SLAM utilizes video as the main input to build a 3D map of an unknown environment, which is then used in another processing thread to determine the camera pose within the environment in real-time \cite{cadena2016past}.

The progress in this field has been remarkable over the past three decades, leading to the extensive adoption of SLAM in real-world applications across various industries (e.g., robotics, autonomous driving or augmented and virtual reality). Starting from the introduction of probabilistic formulations for Simultaneous Localization and Mapping (SLAM) \cite{durrant2006simultaneous,bailey2006simultaneous,davison2007monoslam,klein2007parallel}, researchers have investigated the fundamental properties of SLAM, such as observability, convergence, and consistency \cite{civera2008inverse,chli2009active,strasdat2010real,newcombe2011dtam,dissanayake2011review,kerl2013robust}. Current pipelines \cite{engel2014lsd,mur2015orb,forster2016svo, mur2017orb, engel2017direct,schops2019bad,teed2021droid} show an impressive performance in terms of accuracy and robustness. However, although successful implementations in certain domains might rise the question \textit{``is fundamental research in SLAM still necessary?''}, challenges still remain, particularly in complex and dynamic environments \cite{bescos2018dynaslam,xu2019mid,ballester2021dot}. Achieving truly \textbf{robust performance} is critical for a wide range of applications. To ensure that SLAM can provide reliable perception and navigation for long-lived autonomous robots, fundamental aspects such as \textit{self-tuning}, \textit{resource awareness}, and \textit{task-driven perception} are yet to be addressed \cite{cadena2016past}.

% However, challenges still remain, particularly in dealing with complex and dynamic environments \cite{bescos2018dynaslam,xu2019mid,ballester2021dot}. Ongoing research, what has been called the \textit{robust-perception age of SLAM} \cite{cadena2016past}, seeks to improve the performance and scalability of SLAM for even more demanding applications, in particular safety-critical ones.

%However, with the advancements in computer vision, machine learning, and sensor technologies, it raises the question: \textit{is fundamental research in SLAM still necessary?} While SLAM has made significant progress over the past few decades, achieving truly \textbf{robust performance} remains critical for a wide range of applications that require SLAM with globally consistent mapping. To ensure that SLAM can provide reliable perception and navigation for long-lived autonomous robots, future research efforts should focus on fundamental tasks such as \textit{self-tuning capabilities}, \textit{resource awareness}, and \textit{task-driven perception} \cite{cadena2016past}.

\begin{figure}
  \centering
  \subfigure{\label{fig:kitti_globalBias}\includegraphics[width=0.4\textwidth]{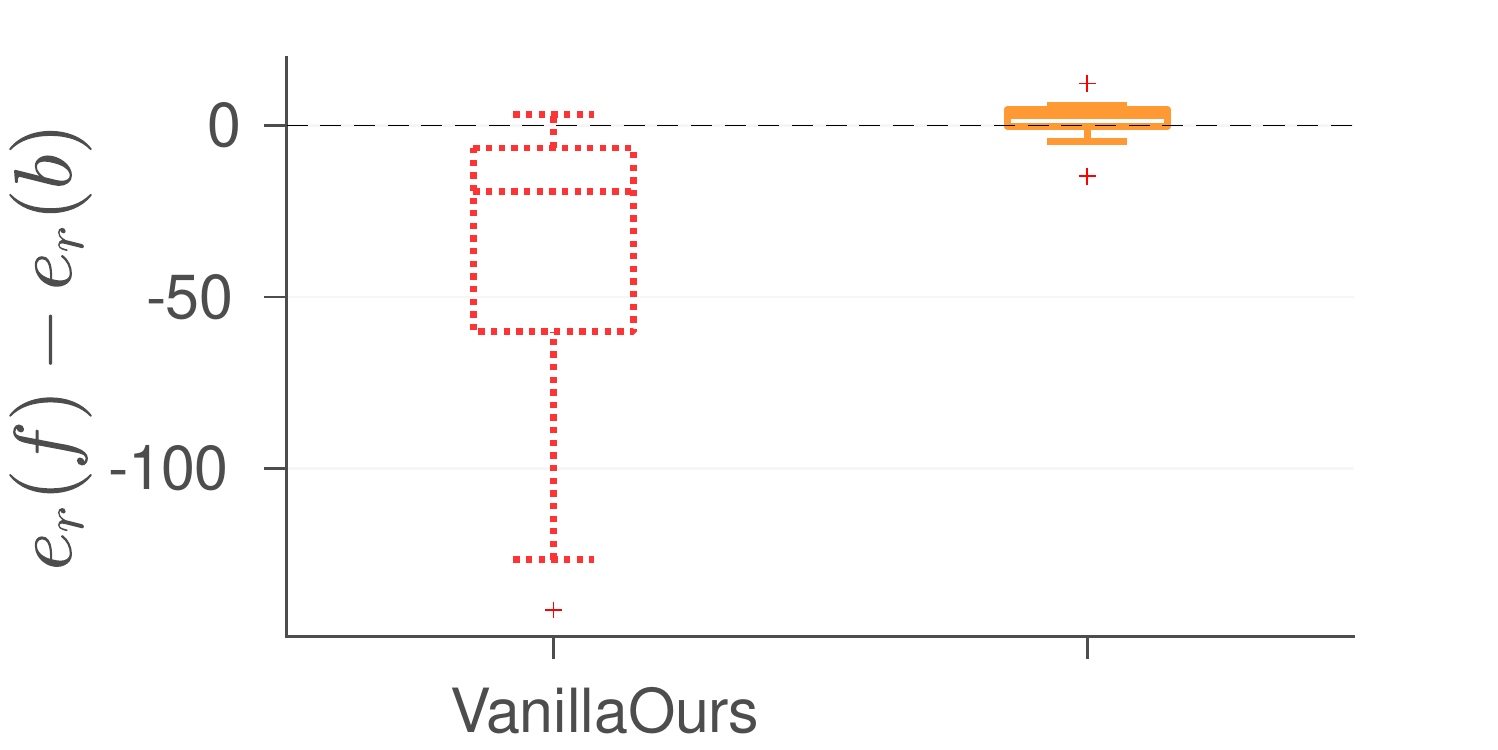}}
  \subfigure{\label{fig:monotum_globalBias}\includegraphics[width=0.4\textwidth]{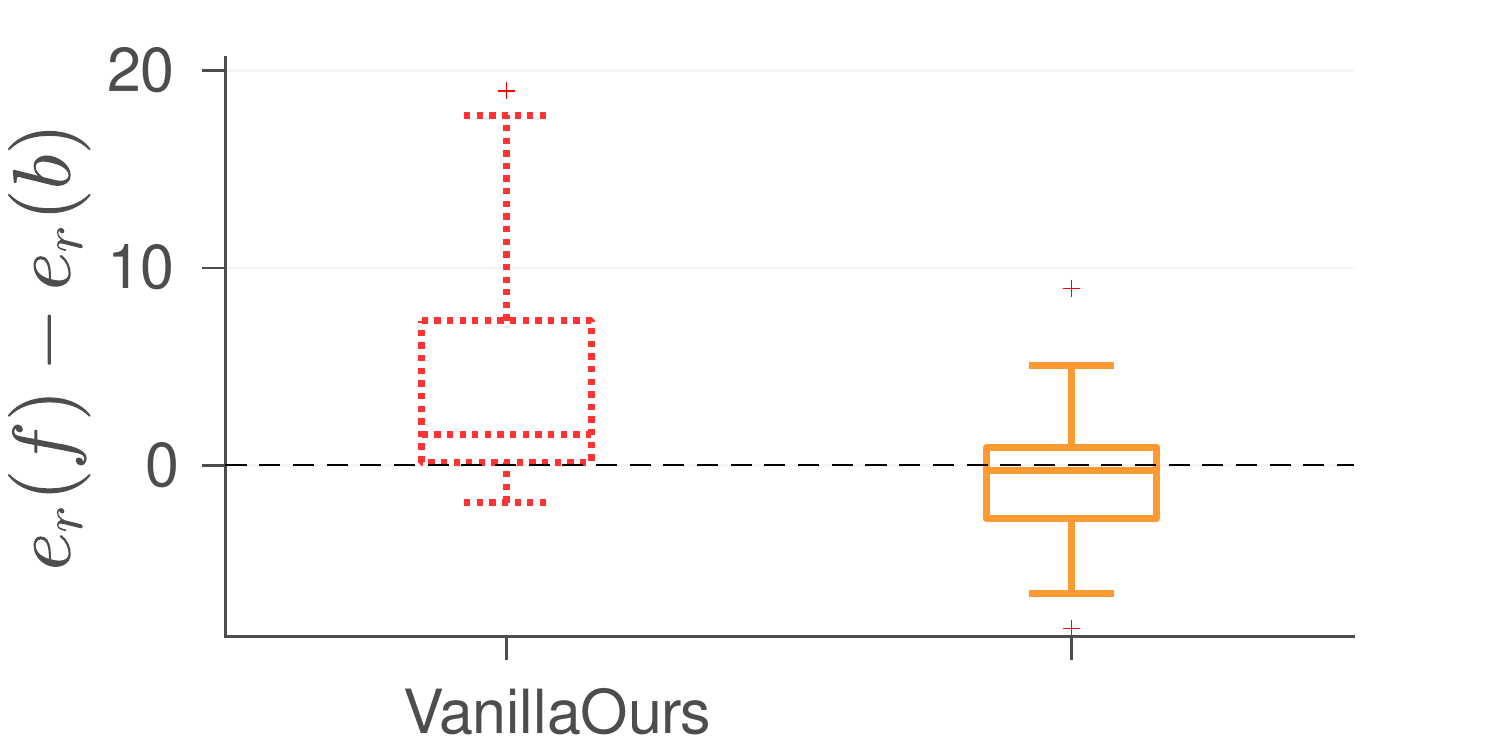}}
  %\setcounter{subfigure}{0}
  %\vspace{-4.3mm}
  \setcounter{subfigure}{0}
  \subfigure[KITTI Dataset (10 sequences)]{\label{fig:kitti_globalAccuracy}\includegraphics[width=0.4\textwidth]{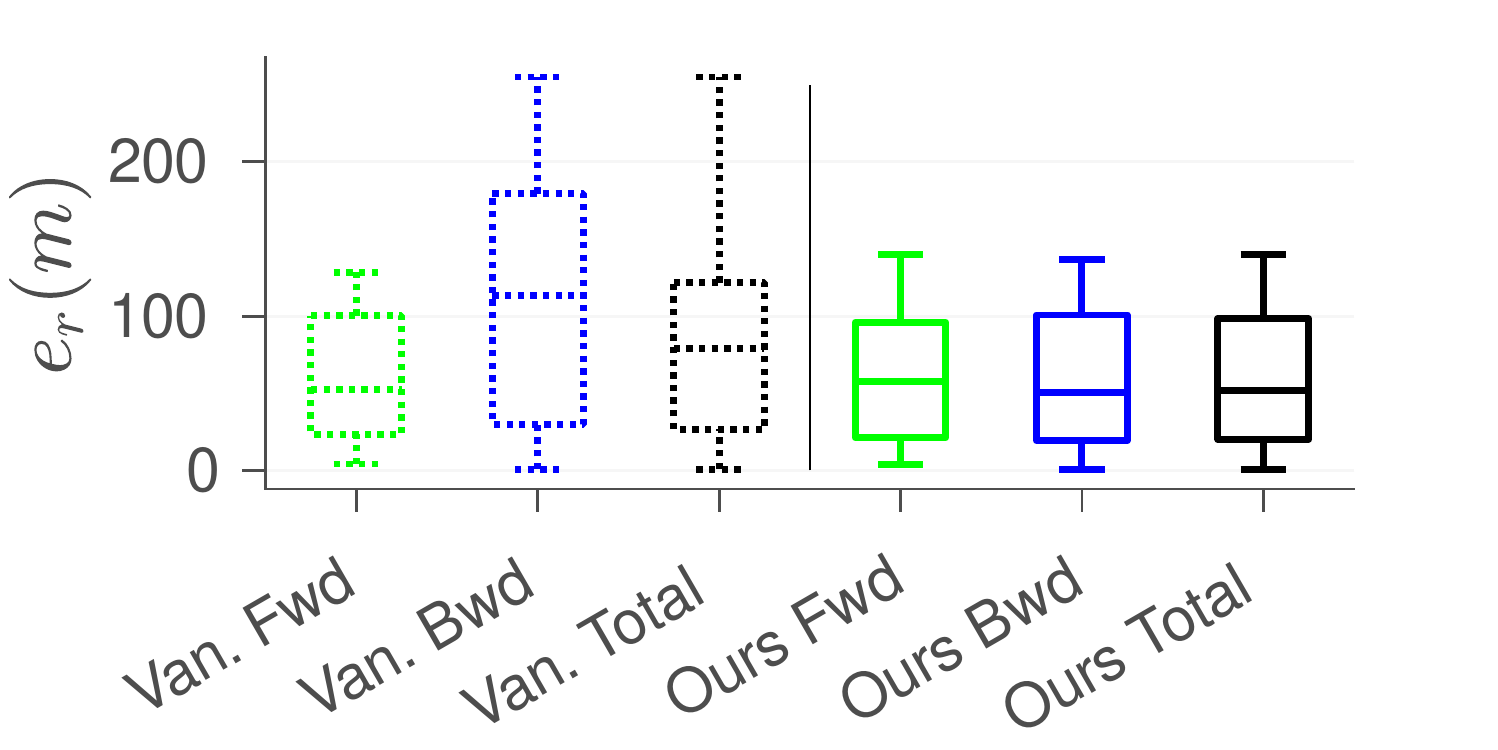}}
  \subfigure[TUM Mono VO Dataset (50 sequences)]
  {\label{fig:monotum_globalAccuracy}\includegraphics[width=0.4\textwidth]{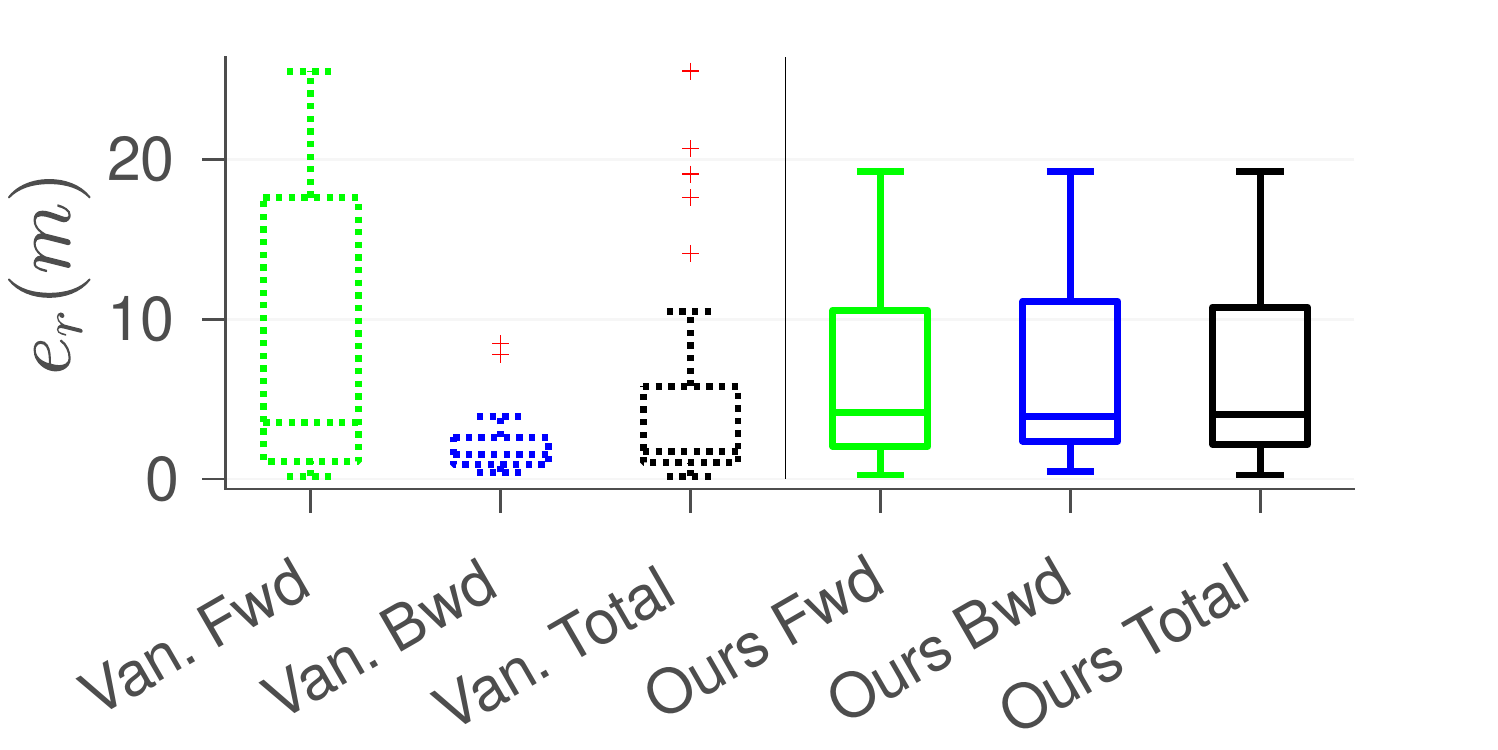}}

  \caption{Motion bias and errors in KITTI \cite{geiger2013vision} \subref{fig:kitti_globalAccuracy} and TUM Mono VO \cite{engel2016photometrically} \subref{fig:monotum_globalAccuracy} datasets. \textbf{Top row displays the motion bias} as the arithmetic error difference between forward (f) and backward (b) passes. Bigger $\pm$ values represent higher motion errors. Note in the box-and-whiskers plots the substantial biases in the Vanilla ORB-SLAM2 version, and how our implementation is able to remove them.
  \textbf{Bottom row shows forward (green), backward (blue) and total (black) errors} for Vanilla ORB-SLAM2 and our version. Note that our errors are similar for forward and backward passes, in contrast to substantial differences in Vanilla ORB-SLAM2. Moreover, observe the opposite bias sign in KITTI and TUM Mono VO, indicating data dependency. 
  }
  \label{fig:vanilla_ORBSLAM}
\end{figure}

In this paper we address \textit{motion bias}, a particular problem that mainly affects feature-based visual SLAM. Several works \cite{engel2016photometrically,engel2017direct,yang2018challenges} and our own results show that variations in the direction of the camera motion (i.e., playing the same image sequence forwards and backwards) lead to significant differences in the performance of SLAM methods. Figure \ref{fig:vanilla_ORBSLAM} contains experimental results showing this bias in KITTI \cite{geiger2013vision} and TUM Mono VO \cite{engel2016photometrically}. 

% The term \textit{motion bias} here specifically refers to the disparity in performance between forward and backward camera movements. Similar to \cite{engel2016photometrically,engel2017direct,yang2018challenges}, we evaluate \textit{motion bias} by running a set of sequences forward and backward.

%We refer as \textit{consistency} to the capacity of a SLAM baseline to keep the accuracy of its estimations within the boundaries of uncertainty. 

The existence of this motion bias in SLAM pipelines is relevant and should be addressed. On the one hand, motion bias should not appear from an optimization perspective, as the geometric configuration and scene appearance do not change when playing an image sequence forwards or backwards. This reveals flaws in current feature-based pipelines, and lack of explainability of their inner working modes. It may also indicate that they are tuned to overfit benchmarks and reported results do not generalize in more general setups.

On the other hand, bias-free forward and backward estimates are desirable for SLAM to generalize across a wide range of use cases. We can cite several application domains in which agents perform varied motions, such as robots equipped with a front-facing and a rear-facing camera \cite{agarwal2020ford,maddern20171,meyer2021madmax}, highly maneuverable robots such as vacuum cleaners, drones or robotic arms, and systems operating in environments with strong motion constraints, such as mine exploration or endoscopic surgery \cite{recasens2021endo}. 

It is true that there might be applications with dominant motion patterns, such as autonomous driving. However, it should be remarked that the motion bias in current SLAM pipelines is neither consistent nor predictable, impacting in their reliability and trustworthiness in critical applications. Figure \ref{fig:vanilla_ORBSLAM} shows how ORB-SLAM2 presents opposite biases 
in KITTI and TUM Mono VO: forward motion estimates are more accurate in KITTI, whereas backward ones are more accurate in TUM Mono VO. 

In this paper we make the following contributions: (1) we extend previous evaluations \cite{engel2016photometrically,engel2017direct,yang2018challenges} that characterized the bias present in feature-based SLAM for forward and backward camera motion, (2) we identify potential sources that may cause such bias and provide theoretical insights and empirical analysis for a better design of new methods, (3) we implement several novel contributions for motion bias removal on the state-of-the-art ORB-SLAM2 \cite{mur2017orb}, specifically in a deterministic version\footnote{Our deterministic implementation of ORB-SLAM2 \cite{mur2017orb} described in Section \ref{sec:ablationStudies} can be found in \href{https://github.com/alejandrofontan/ORB_SLAM2_Deterministic.git}{https://github.com/alejandrofontan/ORB\_SLAM2\_Deterministic}.} to facilitate ablation studies and analyses, and (4) we evaluate our proposals in 4 different and complementary datasets. Our results show that we are able to reduce drastically the motion bias in feature-based SLAM, while keeping the general accuracy.

\section{Related Work}
\label{sec:relatedWork}
Motion Bias was unveiled by Engel et al. \cite{engel2017direct}, that showed the difference in accuracy of feature-based monocular ORB-SLAM2 \cite{mur2017orb} between forward and backward runs in the TUM Mono VO Dataset \cite{engel2016photometrically}. They arrived at three significant outcomes: 1) They showed that ORB-SLAM2 performed significantly better for backwards motion, 2) that, contrary to ORB-SLAM2, their direct sparse odometry (DSO) was largely unaffected by motion bias, and 3) that image resolution was a relevant factor for ORB-SLAM2, while DSO was only marginally affected due to subpixel accuracy. From these outcomes they also concluded that benchmarking on large datasets, covering a diverse range of environments and motion patterns, was of high importance. While Engel et al. raised these issues, no analysis, conclusion, or possible remedies were given in their work. 

Yang et al. \cite{yang2018challenges} performed systematic and quantitative evaluations of motion bias on the three most popular SLAM formulations, namely direct, feature-based and semi-direct methods. Their initial findings agreed with those of Engel et al. \cite{engel2017direct}. They claim that the performance degradation of feature-based SLAM come from implementation details, and proposed that future work should take into consideration the following issues: \textit{depth representation}, \textit{point sampling strategy}, \textit{point management} and \textit{discretization artifacts}. They implemented a version of ORB-SLAM2 with sub-pixel matching accuracy that reduced the errors of the forward passes but maintained a substantial gap between forward and backward motions. Motion bias remained hence unsolved, and Yang et al.'s final conclusion highlighted the importance of the problem, stating that more efforts should be taken to address such problem for the adoption of feature-based SLAM in safety-critical applications like autonomous driving. In this paper we continue their work by formulating and evaluating six novel contributions that realistically minimize motion bias with minimal impact in SLAM general accuracy.

%\begin{figure}
%  \centering
%  \subfigure{\label{fig:subfig1}\includegraphics[width=0.40\textwidth]{figures/LOOP_imageOrb_crop.pdf}
%  \includegraphics[width=0.022\textwidth]{figures/LOOP_imageLegend_crop.pdf}
%  }
%  \subfigure{\label{fig:subfig2}
%  \includegraphics[trim=1cm 6cm 6.5cm 0cm,clip,width=0.5\textwidth]{figures/Fig8.pdf}}
%  \subfigure[Caption for subfigure 2]{\label{fig:subfig2}\includegraphics[width=0.4\textwidth]{figures/LOOP_imageDSVO_crop.pdf}\includegraphics[width=0.022\textwidth]{figures/LOOP_imageLegend_crop.pdf}}
%  \subfigure[Caption for subfigure 2]{\label{fig:subfig2}\includegraphics[trim=1cm 0cm 6.5cm 6cm,clip,width=0.5\textwidth]{figures/Fig8.pdf}}
%  \caption{THIS FIGURE COLLECTS THE RESULTS FROM \cite{engel2016photometrically,engel2017direct,yang2018challenges} I'M NOT SURE IF I WILL KEEP THIS FIGURE???}
%  \label{fig:relatedWork}
%\end{figure}

\section{Motion-Bias-Free Feature-Based SLAM}
\label{sec:motionBias}
In this section, we formulate the contribution of visual covariances to motion bias and introduce an alternative robust cost function for outliers. We also address the sources of the motion bias by providing theoretical insights and proposing motion-bias-free models for five components within feature-based SLAM, specifically related to \textit{point representation}, \textit{optimization} and \textit{data association}.

\subsection{Point Representation}
\label{sec:pointRepresentation}
Feature-based SLAM estimates a sparse map of 3D points $\{ \textbf{p} \in \mathcal{P} \mid \textbf{p} \in \mathbb{R}^3 \}$ and keyframes $k \in \mathcal{K}$. Keyframes are composed by a grayscale image $\textbf{I}_k$ and its 6DOF camera pose, represented by a rotation matrix $\textbf{R}_k \in \mathrm{SO}(3)$ and translation vector $\textbf{t}_k \in \mathbb{R}^3$. Each point $\textbf{p}$ is associated with a set of 2D keypoints $\mathcal{U} \equiv \{\mathbf{u}_{k1}, \mathbf{u}_{k2}, \dots, \mathbf{u}_{kn}\}, \mathbf{u}_{ki} \in \boldsymbol{\Omega}_{ki}$ with corresponding feature descriptors $\mathcal{B} \equiv \{\mathbf{b}_{k1}, \mathbf{b}_{k2}, \dots, \mathbf{b}_{kn}\}, \mathbf{b}_{ki} \in \mathbb{R}^d$. The keypoints correspond to the projections of $\textbf{p}$ in the keyframes $\mathcal{K}$ in which it is visible, and $\boldsymbol{\Omega}_{ki}$ are their image domains. 

\textbf{Reference Descriptor Selection.} To enhance the matching process and acquire more observations with new keypoints, a common technique is to choose a distinctive reference descriptor $\mathbf{b}_r$ and match subsequent keypoints against it \cite{mur2015orb, mur2017orb, campos2021orb}. Traditionally, the reference descriptor is selected based on image appearance. For instance, one approach is to choose the descriptor with the least median Hamming distance $d_H$ to the remaining descriptors: $\mathbf{b}_r = \operatorname{argmin}_{\mathbf{b}_{ki}} \operatorname{median}(d_H(\mathbf{b}_{ki},\mathcal{B})))$ \cite{mur2017orb}. In contrast, we propose to leverage 3D geometry by selecting the descriptor with the smallest Euclidean distance $d_E$ between the translation vectors of its holding keyframe $\textbf{t}_k$ and the keyframe $\textbf{t}_i$ containing the keypoint being matched: $\mathbf{b}_r = \operatorname{argmin}_{\mathbf{b}_k} d_E(\textbf{t}_k,\textbf{t}_i))$. With this policy, we aim to enhance matching quality by utilizing the descriptor that exhibits closest geometric proximity.

\textbf{Point Appearance Invariance.} In order to reduce the number of incorrect observations resulting from the matching process, we apply a geometric constraint that is robust to motion bias to filter the query keypoints $u_q$. The image keypoints $\mathcal{U}$ are detected at different octaves $l \in (0,l_{max} \in \mathbb{Z})$ corresponding to scaled images, where $w(l) = s^l \cdot w_0$ and $h(l) = s^l \cdot h_0$. Here, $s$ represents the scaling factor, and $w_0$ and $h_0$ are the widths and heights at the highest resolution. To establish the appearance invariance interval for the depth $z_q$ of the query keypoints $u_q$, we set a maximum change in the octave of the keypoints denoted by $\Delta l$:
\begin{equation}
   \small{z_q \in [\operatorname{max}_k(z_k \cdot s^{-\Delta l - 0.5}) , \quad \operatorname{min}_k(z_k \cdot s^{\Delta l + 0.5})],}
   \label{eq:depth}
\end{equation}
where  $z_k$ is the depth of the point $\textbf{p}$ with respect to the keyframe $k$ (see Figure \ref{fig:depthInv}).

\subsection{Optimization}
\label{sec:optimizationProcess}

\textbf{Visual Covariances contribution to motion bias.} The camera pose optimization in feature-based SLAM is based on minimizing a weighted residual $\textbf{r}\boldsymbol{\sigma}^{-2}_r\textbf{r}^\top$, in which $\textbf{r}$ and $\boldsymbol{\sigma}^2_r$ are respectively the reprojection error and its covariance. In this section we formulate the contribution to motion bias of the standard approximation of visual covariances and propose a new cost function that is robust to motion bias.

We identify an \textit{observation} as the projection of a 3D point $\textbf{p}$ in two different keyframes $\{i,j\} \in \mathcal{K}$. We label with subscript $j$ the reference frame, and with $i$ any other frame from which the point is visible. The image coordinates of the projection of \textbf{p} in reference frame $j$ and its depth are denoted as $\textbf{u} \in \boldsymbol{\Omega}$ and $z \in \mathbb{R}$ respectively (we drop subindex $j$ for simplicity).

The function $\varphi(\textbf{u})$ projects a point $\textbf{p}$ from its camera coordinates $\textbf{u}$ in the reference frame $j$ into the frame $i$,

\begin{equation}
\small{\varphi(\textbf{u}) = \Pi(\textbf{R}\Pi^{-1}(\textbf{u},z) + \textbf{t}),}
\label{eq:projFunct}
\end{equation}

\noindent where $\Pi(\textbf{p})$ (determined by the intrinsic camera parameters) projects $\textbf{p}$ in the image; and $\Pi^{-1}(\textbf{u},z)$ back-projects the image point with coordinates $\bf{u}$ at depth $z$. $\textbf{R} \in \mathrm{SO}(3)$ and $\textbf{t} \in \mathbb{R}^3$ stand here for the relative rotation and translation between keyframes $j$ and $i$.

Feature-based errors are usually defined as variations of the following expression:

\begin{equation}
\small{\textbf{r} = \textbf{u}_i-\varphi(\textbf{u}) , \quad \boldsymbol{\sigma}_r^2 =  \boldsymbol{\sigma}^2_{u_i} +  \varphi( \boldsymbol{\sigma}^2_{u}),}
\label{eq:residual}
\end{equation}

where $\textbf{u}_i$ $\{ \textbf{u}_i \in \mathcal{U} \mid \textbf{u}_i \in \mathbb{R}^2 \}$ stands for the feature point in image $i$ and $\varphi(\textbf{u})$ for the corresponding point in the frame $j$ reprojected in the frame $i$. Hence the residual covariance is expressed as the sum of the projection covariance and the feature subpixel noise $\boldsymbol{\sigma}_{u_i}^2$.

We approximate the residual function \eqref{eq:residual} by its \textbf{first-order Taylor approximation} 
\begin{equation}
\small{\varphi(\textbf{u}+ d\textbf{u}) \approx \varphi(\textbf{u}) + \nabla_{\textbf{u}} \varphi d\textbf{u} , \quad \boldsymbol{\sigma}^2_r \approx \boldsymbol{\sigma}^2_{u_i} + \nabla_{\textbf{u}} \varphi\boldsymbol{\sigma}^2_{u} \nabla_{\textbf{u}} \varphi^T ,}
\end{equation}
with the \textbf{perspective deformation gradient tensor} $\nabla_{\textbf{u}} \varphi$ \cite{fontan2022model} being the Jacobian of the projection function \eqref{eq:projFunct}.

Assuming pure forward/backward movement, $\textbf{R} = \textbf{I}_3$, $\textbf{t} = (0, 0, t)^T$, the feature subpixel noise,  the gradient tensor and the residual covariances can be modeled as isotropic Gaussians, 

\begin{equation}
    \small{\boldsymbol{\sigma}^2_r = \sigma^2_r\textbf{I}_2 = (\sigma^2_{u_i} + \varepsilon_{ij}^2\sigma^2_{u_j})\textbf{I}_2 \quad ,\text{with} \quad \boldsymbol{\sigma}^2_{u_i} = \sigma_{u_i}^2 \textbf{I}_2  \quad , \boldsymbol{\sigma}^2_{u_j} = \sigma_{u_j}^2 \textbf{I}_2 \quad ,\text{and} \quad  \nabla_{\textbf{u}} \varphi = \varepsilon_{ij} \textbf{I}_2.}
    \label{eq:perspectiveDeformation}
\end{equation}

The \textbf{standard approach for visual covariance }is to approximate the residual covariance of the keypoint as $\sigma^2_r|_{appr} = 2\sigma^2_{u_i}$ \cite{mur2015orb}. Compared against the value from the linearization

\begin{equation}
    \small{\alpha = \frac{\sigma^2_r|_{appr}}{\sigma^2_r} =  
    \frac{2\sigma^2_{u_i}}{(\sigma^2_{u_i} +  \varepsilon_{ij}^2 \sigma^2_{u_j})} =  
    \frac{2}{(1 + \varepsilon_{ij}^2)} \left\{
\begin{array}{c l}
 > 1&  \text{if \hspace{2mm}} \varepsilon_{ij}^2 < 1 \text{\hspace{2mm} \textit{ , over-estimation} }\\
  & \\
 < 1  & \text{if \hspace{2mm}} \varepsilon_{ij}^2 > 1 \text{\hspace{2mm} \textit{ , under-estimation.} }
\end{array}
\right.}
    \label{eq:alpha_vanilla}
\end{equation}

\noindent this leads to over- and under-estimation effects that result in different outlier rejection criteria at forward and backward motions. To address this issue, we compute the reprojection error in both the projection and reference keyframes (assuming constant depth) and normalize them with the covariance of the keypoints in their respective views

\begin{equation} \label{eq:geoCostFunction}
        \small{\textbf{r}_i = (\textbf{u}_i-\varphi(\textbf{u}))^2 ,\textbf{\hspace{4mm}} \textbf{r}_j = (\textbf{u}-\varphi(\textbf{u}_i))^2
        ,\textbf{\hspace{4mm}}\boldsymbol{\sigma}_{r_i}^2 =  2\boldsymbol{\sigma}^2_{u_i}
        ,\textbf{\hspace{4mm}}\boldsymbol{\sigma}_{r_j}^2 =  2\boldsymbol{\sigma}^2_{u_j} }.
\end{equation} 

\begin{figure}
  \centering
  \subfigure[\textbf{Depth invariance interval.}]{\label{fig:depthInv}\includegraphics[width=0.39\textwidth]{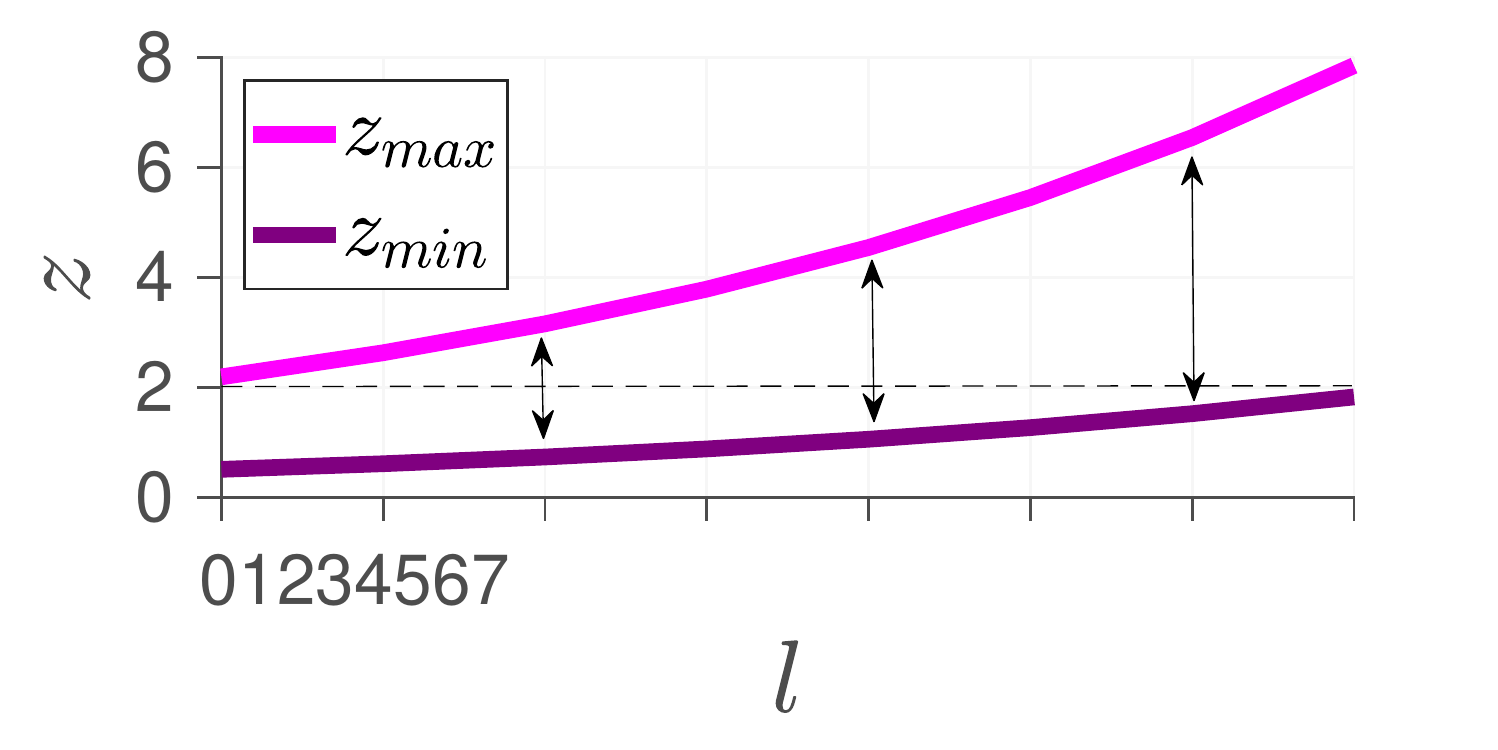}}
  \subfigure[\textbf{Covariance over-/under-estimation.}]{\label{fig:covarianceModel}\includegraphics[width=0.39\textwidth]{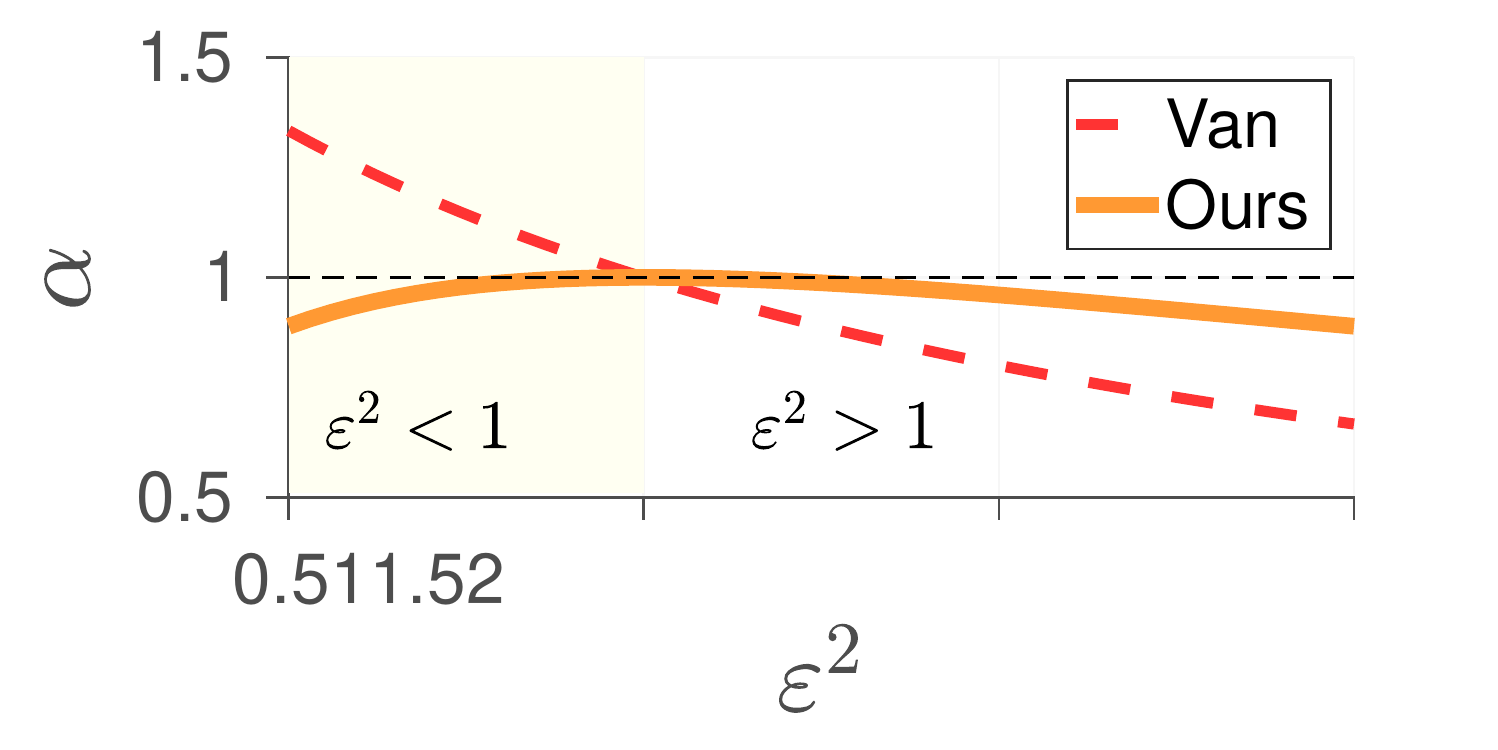}}
  \caption{\textbf{Left:} We show the \textbf{depth invariance interval} \eqref{eq:depth} for every octave of a point found at $z = 2 (m)$. \textbf{Right:} Relative error in the estimation of visual covariances \eqref{eq:alpha_vanilla}, \eqref{eq:alpha_ours}.}
\end{figure}

Figure \ref{fig:covarianceModel} compares the robustness of the standard approach and our reprojection function against motion bias
\begin{equation}
    \small{
    \alpha = \frac{\sigma^2_r|_{appr}}{\sigma^2_r} =  
    \frac{2\sigma^2_{u_i} + 2\sigma^2_{u_j}}{(1 + \varepsilon_{ij}^2)\sigma^2_{u_j} +  (1 + \varepsilon_{ji}^2)\sigma^2_{u_i}} \stackrel{\sigma^2_{u_i} = \sigma^2_{u_j}}{=}  
    \frac{4}{2 + \varepsilon_{ij}^2 + \varepsilon_{ji}^2}.}
    \label{eq:alpha_ours}
\end{equation}

\textbf{Outlier Rejection.} ORB-SLAM2 \cite{mur2017orb} and DSO \cite{engel2017direct}, representative pipelines for feature-based and photometric SLAM, implement two strategies to reject outliers. Firstly, they minimize a Huber cost function \cite{concha2015evaluation} of the geometric and photometric errors respectively in order to downweight their influence. And secondly, both approaches remove point observations that have a Mahalanobis residual above a certain threshold after optimization. Although this second strategy may have benefits, it may have a negative effect if removal is done too early. For instance, for a frame with a large number of noisy observations (due to dynamic objects, motion blur or lighting changes), accurate matches can suffer from early removal due to inaccurate initial estimates.

Early removal of potential outliers may not have a significant influence for photometric methods, as observations are abundant and very efficiently computed by simply retrieving the intensity values of the image points \cite{fontan2023sid}. However, it can degrade the performance of feature-based methods. As features have a high extraction footprint, their re-detection after removal comes with certain delay. In order to avoid this effect, we do not remove potential outliers and keep all observations in the optimization. 

%Classifying observations as outliers is subject to several factors, including the uncertainty of the current system state, the residual threshold set, and for feature-based approaches, the quality of the data association.

\subsection{Data Association}
\label{sec:dataAssociation}

\begin{figure}

\begin{minipage}{0.30\linewidth}
\begin{algorithm}[H]\scriptsize
%\caption{\scriptsize \newline ORB-SLAM2 Tracking.}
\begin{algorithmic}[1]
    \Function{Tracking }{$\mathcal{I}$}
    \State \Comment{$\mathcal{I}$ = RGB image}
    \If{Not Initialized}
        \State Try to initialize ($\mathcal{I}$)
    \EndIf
    \State Track With Motion Model ()
    
    \hspace{0.1cm} {\fontsize{6.9}{6}\selectfont \underline{Search By Projection v1 ($c_1$)}} 
    \State Track Local Map ()
    
    \hspace{0.1cm} {\fontsize{6.9}{6}\selectfont \underline{Search By Projection v2 ($c_2$)}}
    \If{Need New KeyFrame ()}
        \Let {$\mathcal{K}_n$}{\fontsize{5.7}{4}\selectfont Create New KeyFrame () \scriptsize}
    \EndIf
    
    \EndFunction
\end{algorithmic}
\end{algorithm}
\end{minipage}
\hfill
\hspace{-1.0cm} 
\begin{minipage}{0.30\linewidth}
\begin{algorithm}[H]\scriptsize
%\caption{\scriptsize \newline ORB-SLAM2 Mapping.}
\begin{algorithmic}[1]
    \Function{Mapping }{$\mathcal{K}_n$}
    \State \Comment{$\mathcal{K}_n$ = new keyframe}
    
    \Let {$\mathcal{P}$}{Map Point Culling ($\mathcal{P}$)}
    
    \Let {$\mathcal{P}_n$}{{\fontsize{6.5}{6}\selectfont   Create New Pts} ($\mathcal{K}_n$,$\mathcal{K}$)}
    
    \hspace{0.5cm} {\fontsize{5.8}{6}\selectfont \underline{Search For Triangulation ($c_3$)}} 

    \State Search In Neighbors ()

    \hspace{0.5cm} \underline{Fuse ($\mathcal{P}_n$,$\mathcal{K}$,$c_4$)}
    
    \hspace{0.5cm} \underline{Fuse  ($\mathcal{P}$,$\mathcal{K}_n$,$c_4$)}

    \State $\mathcal{P} \leftarrow \mathcal{P} \cup \mathcal{P}_n$ , $\mathcal{K} \leftarrow \mathcal{K} \cup \mathcal{K}_n$
    
    \State {\fontsize{6.6}{6}\selectfont Local Bundle Adjustment} ($\mathcal{P}$,$\mathcal{K}$)   
    
    \hspace{0.5cm} Erase Outliers ()
    
    \Let {$\mathcal{K}$}{KeyFrame Culling ($\mathcal{K}$)}
    
    \EndFunction
\end{algorithmic}
\end{algorithm}
\end{minipage}
\hfill
\begin{minipage}{0.25\linewidth}
\hspace{-0.5cm}
\begin{minipage}{1.0\linewidth}
\subfigure{\label{fig:subfig1a}\includegraphics[width=1.0\textwidth]{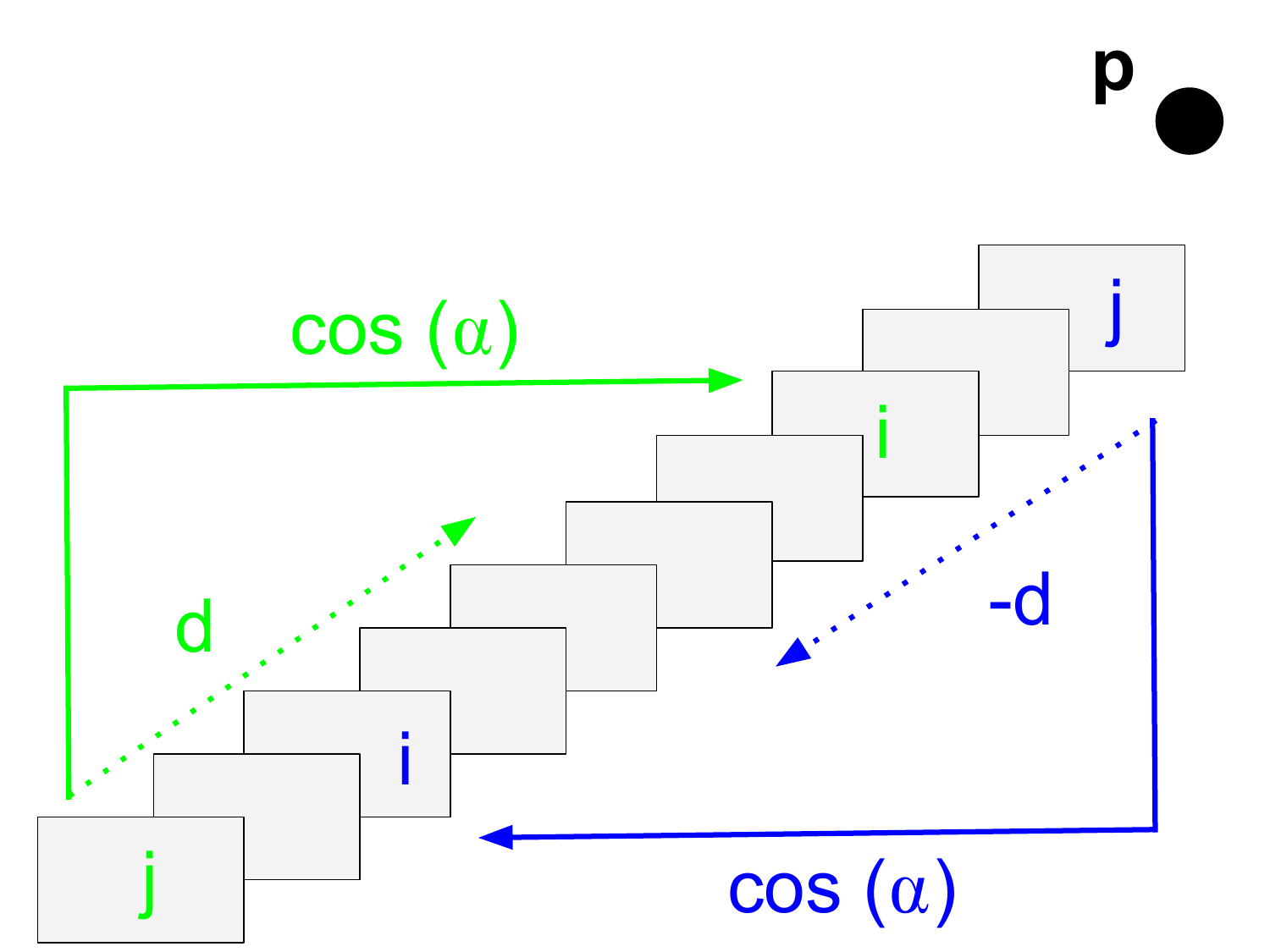}}
\end{minipage}

\vspace{0.3cm}

\hspace{-1.0cm}
\begin{minipage}{1.1\linewidth}
\subfigure{\label{fig:subfig1b}\includegraphics[width=1.13\textwidth]{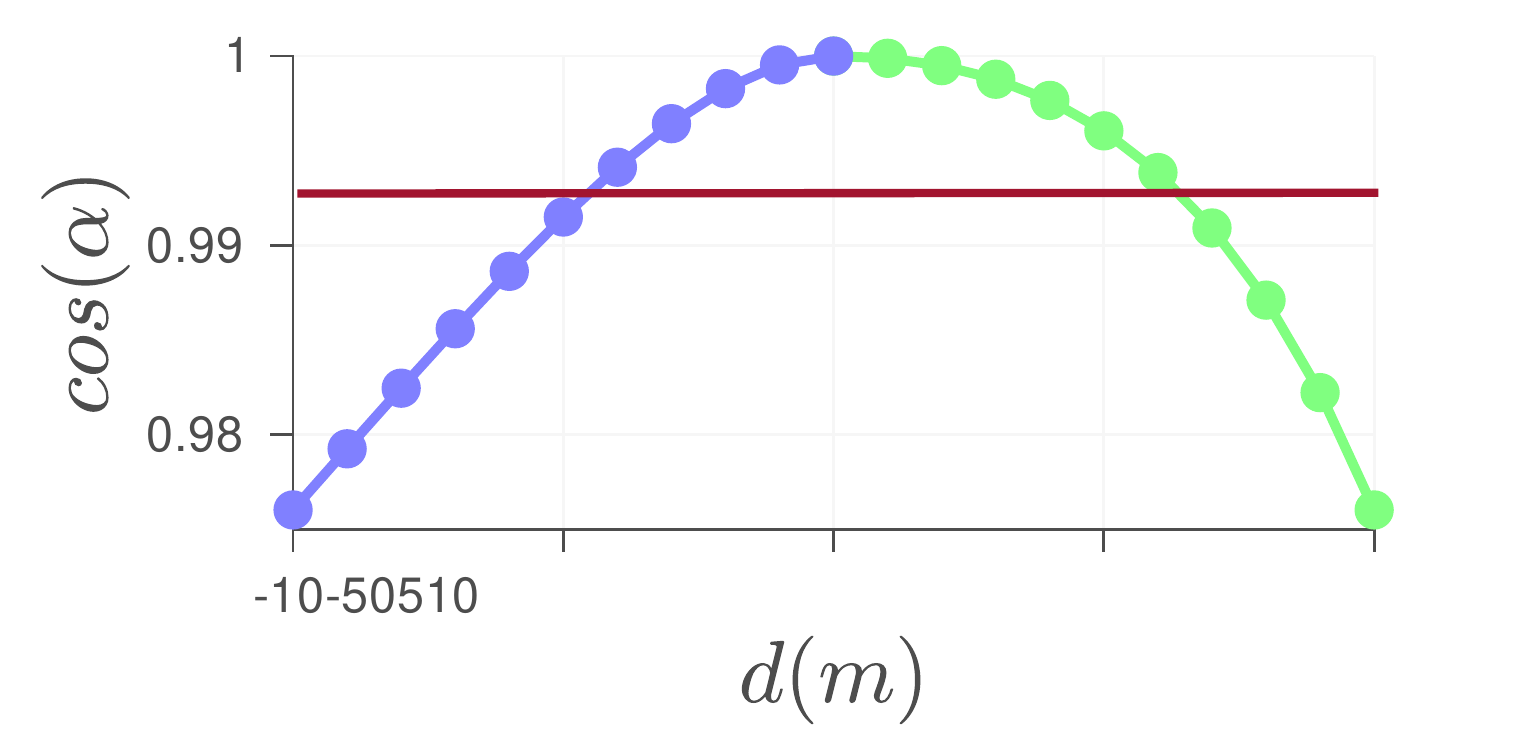}}
\end{minipage}
\end{minipage}
%\hspace{1.2cm}
\vspace{0.2cm}
\caption{\textbf{Left:} \textbf{Tracking and mapping threads} of  \cite{mur2017orb} illustrating the distribution of data association functions throughout a standard pipeline (underlined in the pseudo-code). $c_{1,2,3,4}$ represents variations in the chosen descriptor distance thresholds used for feature matching. \textbf{Right:} \textbf{Asymmetry in the angle} between observations of a point under fwd/bwd motion. }
\label{fig:symmetricMatching}
\end{figure}

\textbf{Symmetric Data Association.} The implementation of data association varies depending on its specific function within the SLAM pipeline. The left part of Figure \ref{fig:symmetricMatching} provides pseudo-code for the tracking and mapping threads of ORB-SLAM2 \cite{mur2017orb}. Design choices differ based on whether the association aims to create new map points (i.e., \textit{Search For Triangulation}), search for new observations (i.e., \textit{Search By Projection}), or fuse map points (i.e., \textit{Fuse}).

The ray from a point observation $\textbf{u}_i \in \mathbb{R}^{3}$ on an image $I \in \{i,j\}$ is $\textbf{r}_i = R^{wc}_i(K^{-1}\textbf{u}_{i})$, where $K^{-1} \in \mathbb{R}^{3\times3}$ contains the camera intrinsics and $R_i^{wc}$ the rotation matrix of the corresponding camera pose. The parallax angle between two rays is $ \cos(\alpha) = (\textbf{r}_i \cdot \textbf{r}_j)/(|\textbf{r}_i||\textbf{r}_j|)$. Figure \ref{fig:symmetricMatching} shows the angle between point observations. Despite the angle at the end of both forward and backward passes being the same, it varies throughout the trajectory. Consequently, if a threshold for the angle is set during point initialization, the number of past and future observations may differ between the forward and backward passes. This discrepancy becomes more prominent when matching functions utilize distinct parameters or heuristics. These variations can result in a different number of observations during forward and backward motion, ultimately leading to differences in performance. 

In our modification of ORB-SLAM2, we have changed the implementation of association functions to ensure the utilization of consistent geometry and appearance constraints.

\textbf{Robust Data Association}. A common approach to expedite data association is to employ greedy strategies, where the matching of keypoints is performed strictly in sequential order. This sequential ordering can lead to biased matching results, as keypoints may be ordered based on their position in the image or their resolution octave. We have modified the matching strategy in the ORB-SLAM2 code to avoid its potential bias. We prioritize matches based on the lowest Hamming distance instead of relying on sequential ordering. 

\section{Ablation Study}
\label{sec:ablationStudies}
In this section, we evaluate quantitatively how the methods proposed in Section \ref{sec:motionBias} contribute to the reduction of motion bias. Specifically, we examine the effects of \textit{descriptor selection}, \textit{point appearance invariance}, \textit{visual covariances}, \textit{outlier rejection}, \textit{symmetric data association}, and \textit{robust data association}. Following \cite{yang2018challenges}, we evaluate the accuracy of the trajectories by calculating an \textbf{alignment error}, denoted as $e_r$. This error is estimated as the translational root-mean-squared-error (RMSE) between the estimated trajectory, after it is aligned to the start and end segments of the ground truth trajectory. For further details, please refer to \cite{engel2016photometrically}. To quantify \textbf{the motion bias}, we define it as the difference between the alignment error of the forward pass and the alignment error of the backward pass: $ e_r (f) - e_r (b)$.

\subsection{Implementation Details}
\label{sec:implementationDetails}

\textbf{Deterministic ORB-SLAM2.} SLAM systems are susceptible to non-deterministic effects that can significantly impede their performance and hinder our ability to perform ablation studies on specific processes or variables. In this study, we introduce a modified version of ORB-SLAM2 \cite{mur2017orb} that has been specifically designed to guarantee full determinism, rendering it highly suitable for conducting ablation studies\footnote{Our deterministic implementation of ORB-SLAM2 \cite{mur2017orb} maintained all capabilities of the original baseline. However, similar to \cite{engel2016photometrically,yang2018challenges}, we deactivated loop closure to maximize the impact of motion bias in our experiments.}. Specifically, 1) we have altered both the tracking and mapping threads to ensure that they operate sequentially, thereby eliminating online non-deterministic behavior resulting from the real-time implementation. 2) In addition, we have addressed implicit non-deterministic behavior that arose from dynamic memory allocation and  random sample consensus (RANSAC).  

%\textbf{Frame vs Keyframe trajectory.} Keyframe-based SLAM approaches store a subset of representative frames in order to make global bundle adjustment and relocalization feasible in real time. However, evaluation the trajectory accuracy only over those keyframes creates an artificial improvement since these will be the most optimized parts of the estimation. ORB-SLAM2 in its RGB-D and Stereo implementations store the relative position between dropped frames an reference keyframes and recovers the whole trajectory at the end of the sequence. The strong changes produced by the \textit{relaxation} of the trajectory and map after a loop closure in a monocular setup lead to the inconsistency of the relative transformations of the marginalized frames. We enable again the use of those relative position by including them also in the updating process.
 
\textbf{Keyframe management.} Selecting a subset of representative keyframes in order to enable real-time global bundle adjustment and relocalization is an open problem in SLAM \cite{konolige2008frameslam,strasdat2010real,leutenegger2013keyframe,younes2017keyframe,schmuck2019redundancy,fontan2023sid}. State-of-the-art approaches often rely on policies that are designed to maximize performance on specific benchmarks, and may degrade when they are tested on out-of-distribution sequences. To ensure homogeneous executions in our forward/backward experiments across four different datasets, we process every new frame as a keyframe but keep only one out of every five together with the latest five keyframes. These numbers are a reasonable compromise in our experiments between computational efficiency and performance.

\subsection{Ablation Results}
\label{sec:ablation}

Table \ref{tab:AblationStudy} illustrates the impact of each proposed method on accuracy and motion bias. We assess the relative contribution of each method by implementing all of them and then conducting separate evaluations by deactivating each method one at a time.

\textbf{The main factor contributing to motion bias is the outlier rejection}, which creates a significant gap between forward and backward runs by artificially improving one pass while degrading the other. Our contributions effectively balance the forward and reverse passes in both datasets, thereby eliminating motion bias.

\begin{table}[]
\centering
\begin{minipage}{0.6\linewidth}
\resizebox{1.0\columnwidth}{!}{%
\begin{tabular}{l c c c c c c }
& \multicolumn{6}{c}{\textbf{Kitti Dataset} \cite{geiger2013vision} } \\
& \multicolumn{2}{c}{\cellcolor[HTML]{E7FEE9}$e_r (f)$}                                    & \multicolumn{2}{c}{\cellcolor[HTML]{DFE5EF}$e_r (b)$}                                    & \multicolumn{2}{c}{\cellcolor[HTML]{FEF4D2}$e_r (f) - e_r (b)$}                        \\ \cline{2-7} 
 & \scriptsize{rmse} & \scriptsize{mean/std}  & \scriptsize{rmse} & \scriptsize{mean/std} & \scriptsize{rmse} & \scriptsize{mean/std}        \\ \hline
\textbf{Ours (full)}  & 72.7 & 58.3/45.7 &  \textcolor{green}{\uline{\textcolor{black}{{71.4}}}} & 56.8/45.4 & \textcolor{green}{\uline{\textcolor{black}{{6.9}}}} & 1.4/ \textcolor{green}{\uline{\textcolor{black}{{7.1}}}}\\ \hline
{\scriptsize{wo.}} {\small{Descriptor Sel.}} &74.4 &59.4/47.2 &79.0&63.4/49.6&15.1&-3.9/15.3\\ 
{\scriptsize{wo.}} {\small{Point App. Inv.}} &64.7&52.7/39.5&71.9&\textcolor{green}{\uline{\textcolor{black}{{56.7}}}}/46.5&9.4&-4.0/8.9\\ 
{\scriptsize{wo.}} {\small{Robust Assoc.}}  &72.4&57.0/47.1&69.3&57.6/\textcolor{green}{\uline{\textcolor{black}{{40.6}}}}&17.8&\textcolor{green}{\uline{\textcolor{black}{{-0.6}}}}/18.7\\ 
{\scriptsize{wo.}} {\small{Sym. Assoc.}} & \textcolor{red}{\uline{\textcolor{black}{{90.6}}}}&\textcolor{red}{\uline{\textcolor{black}{{72.4}}}}/\textcolor{red}{\uline{\textcolor{black}{{57.3}}}} &71.8&59.1/43.0&24.9&13.3/22.2  \\ 
{\scriptsize{wo.}} {\small{Visual Cov.}} &74.4 & 59.4/47.1 & 72.2 &  59.4/43.5 & 13.2 & 3.0/14.1\\ 
{\scriptsize{wo.}} {\small{Outlier Reject.}}  & \textcolor{green}{\uline{\textcolor{black}{{59.8}}}} & \textcolor{green}{\uline{\textcolor{black}{{50.7}}}}/\textcolor{green}{\uline{\textcolor{black}{{33.4}}}}  & \textcolor{red}{\uline{\textcolor{black}{88.9}}} & \textcolor{red}{\uline{\textcolor{black}{69.7}}}/\textcolor{red}{\uline{\textcolor{black}{58.2}}}& \textcolor{red}{\uline{\textcolor{black}{31.2}}} & \textcolor{red}{\uline{\textcolor{black}{-19.0}}}/\textcolor{red}{\uline{\textcolor{black}{26.1}}}\\ 
 
\end{tabular}
}
\end{minipage}
\begin{minipage}{0.39\linewidth}
\resizebox{1.0\columnwidth}{!}{%
\begin{tabular}{ c c c c c c }
\multicolumn{6}{c}{ \textbf{TUM Mono VO Dataset} \cite{engel2016photometrically} } \\
\multicolumn{2}{c}{\cellcolor[HTML]{E7FEE9}$e_r (f)$}                                    & \multicolumn{2}{c}{\cellcolor[HTML]{DFE5EF}$e_r (b)$}                                    & \multicolumn{2}{c}{\cellcolor[HTML]{FEF4D2}$e_r (f) - e_r (b)$}                        \\ \cline{1-6} 
\scriptsize{rmse} & \scriptsize{mean/std}  & \scriptsize{rmse} & \scriptsize{mean/std} & \scriptsize{rmse} & \scriptsize{mean/std}         \\ \hline
6.7 & 4.8/4.7 &  6.8 & 5.3/4.3 & 3.9& -1.3/3.7\\ \hline
13.0 & 5.9/11.7&5.3&4.3/3.2&5.4&-0.5/5.5\\ 
6.7 &  5.0/4.5 & 6.2 & 4.9/3.7 & \textcolor{green}{\uline{\textcolor{black}{{2.8}}}} & \textcolor{green}{\uline{\textcolor{black}{{-0.1}}}}/\textcolor{green}{\uline{\textcolor{black}{{2.9}}}}\\ 
\textcolor{green}{\uline{\textcolor{black}{{5.8}}}} & 4.5/\textcolor{green}{\uline{\textcolor{black}{{3.8}}}} & \textcolor{red}{\uline{\textcolor{black}{{8.3}}}} & \textcolor{red}{\uline{\textcolor{black}{{5.8}}}}/\textcolor{red}{\uline{\textcolor{black}{{6.0}}}} & 4.3 & -0.5/4.3 \\ 
7.6 & \textcolor{green}{\uline{\textcolor{black}{{4.4}}}}/6.2 & 4.7 & 4.1/2.4 & 4.1 & -1.3/3.9\\ 
8.0 & 5.0/6.2 & 7.0&5.3/4.6 &  6.2 & 0.3/6.3\\ 
\textcolor{red}{\uline{\textcolor{black}{{15.2}}}} & \textcolor{red}{\uline{\textcolor{black}{{7.3}}}}/\textcolor{red}{\uline{\textcolor{black}{{13.5}}}} & \textcolor{green}{\uline{\textcolor{black}{{3.3}}}} & \textcolor{green}{\uline{\textcolor{black}{{2.4}}}}/ \textcolor{green}{\uline{\textcolor{black}{{2.4}}}} & \textcolor{red}{\uline{\textcolor{black}{{6.3}}}} & \textcolor{red}{\uline{\textcolor{black}{{3.3}}}}/\textcolor{red}{\uline{\textcolor{black}{{5.4}}}}\\ 
 
\end{tabular}
}
\end{minipage}
%\caption{\textbf{Ablation Study TUM mono VO dataset}. Every row corresponds to a configuration where all methods except one are implemented.}
\vspace{0.1cm}
\caption{\textbf{Ablation Study in KITTI \cite{geiger2013vision} and TUM Mono VO  \cite{engel2016photometrically}}. We report the root-mean-squared-error (RMSE), mean and standard deviation (std) of the trajectory alignment error ($e_r$) in meters (\textbf{all better if closer to zero}). We show results for the forward (\textit{f}) and backward (\textit{b}) passes, and underline in \textcolor{red}{red}/\textcolor{green}{green} the \textcolor{red}{worst}/\textcolor{green}{best} result for each metric. Values in the last columns evaluate the motion bias effect (\textbf{smaller with errors closer to zero)}. Every row evaluates a configuration where all contributions except one are implemented.}

\label{tab:AblationStudy}
\end{table}

\begin{figure}
  \centering
   \hfill
   \subfigure[\# Map Points]{\label{fig:inliersa}\includegraphics[width=0.3\textwidth]{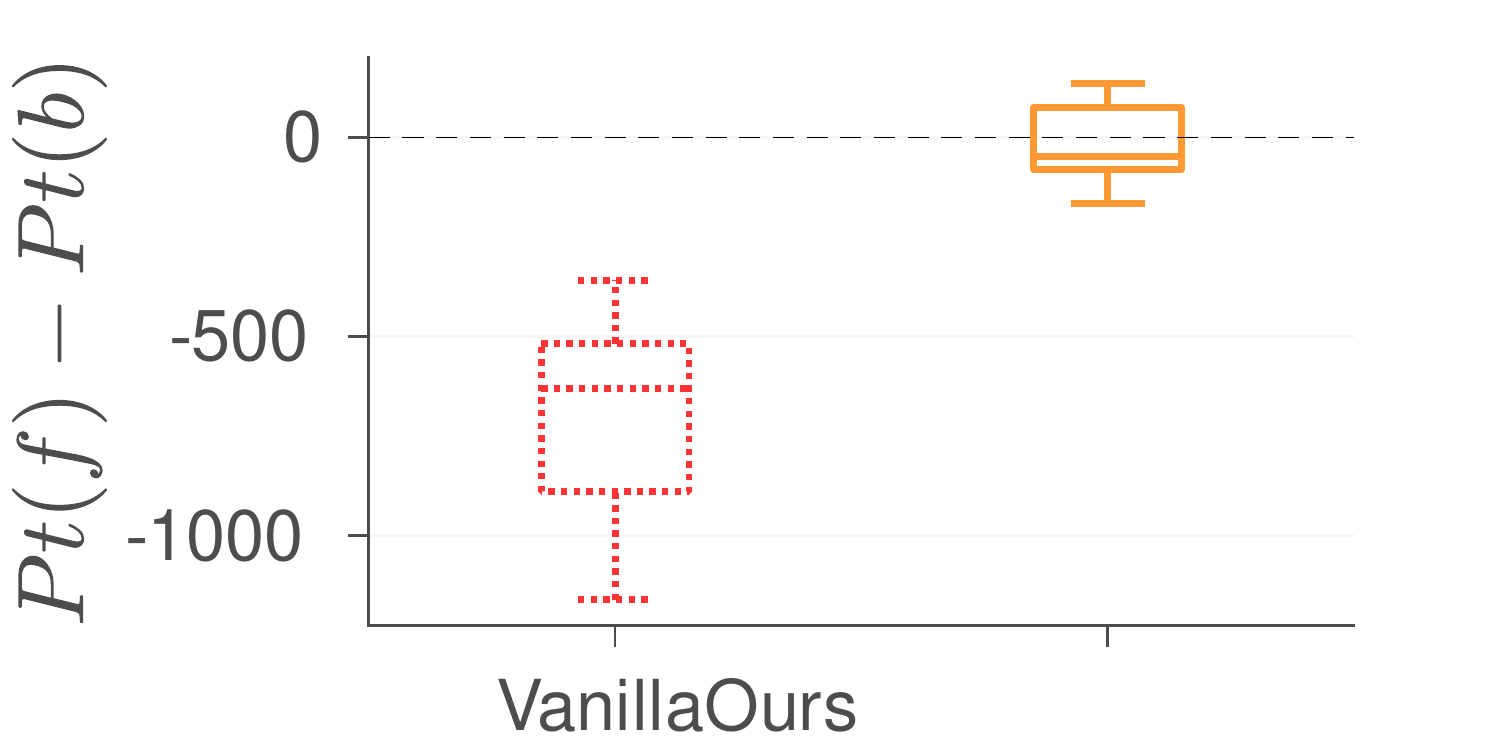}}
   \hfill
   \subfigure[\# Local Keyframes]{\label{fig:inliersb}\includegraphics[width=0.3\textwidth]{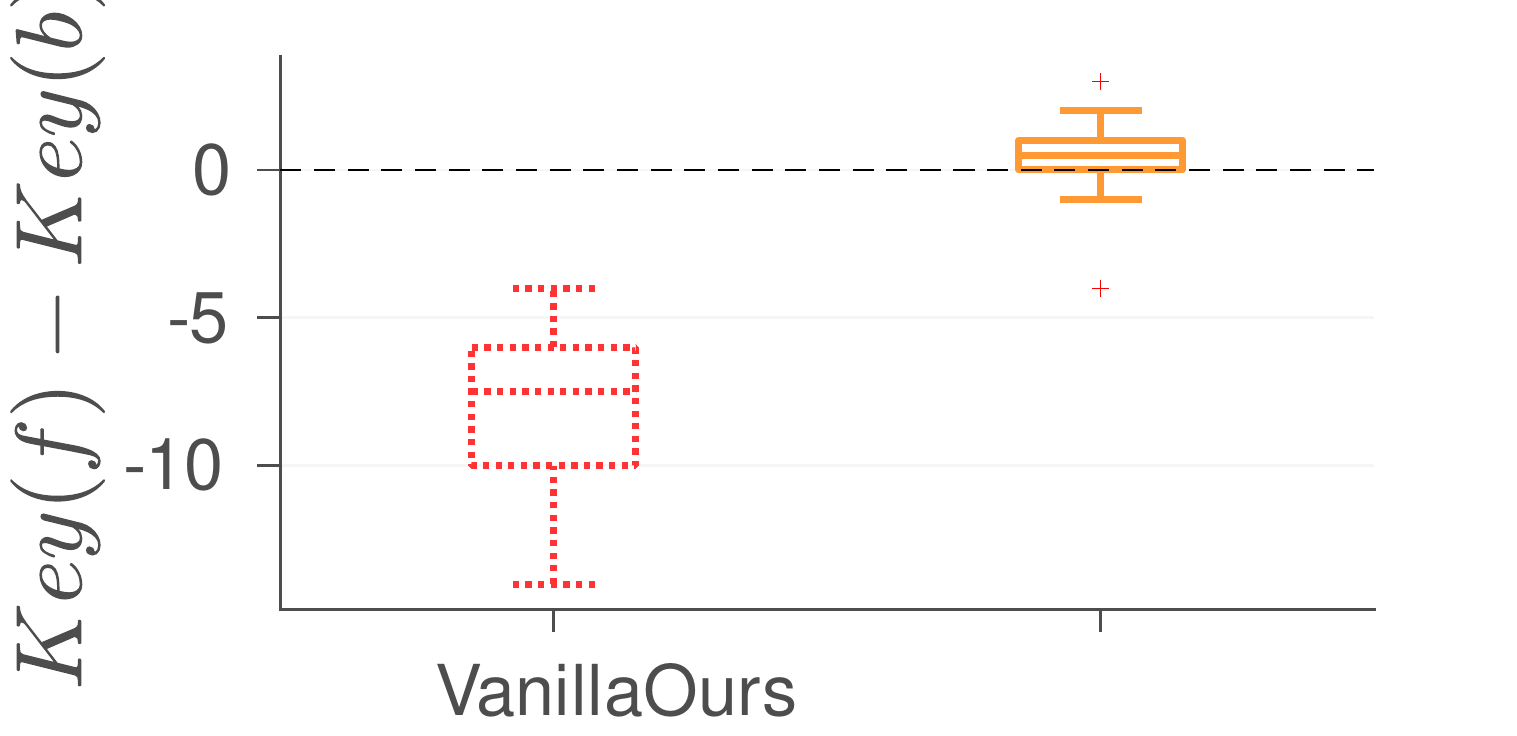}}
   \hfill
   \subfigure[\# Observation Inliers]{\label{fig:inliersc}\includegraphics[width=0.3\textwidth]{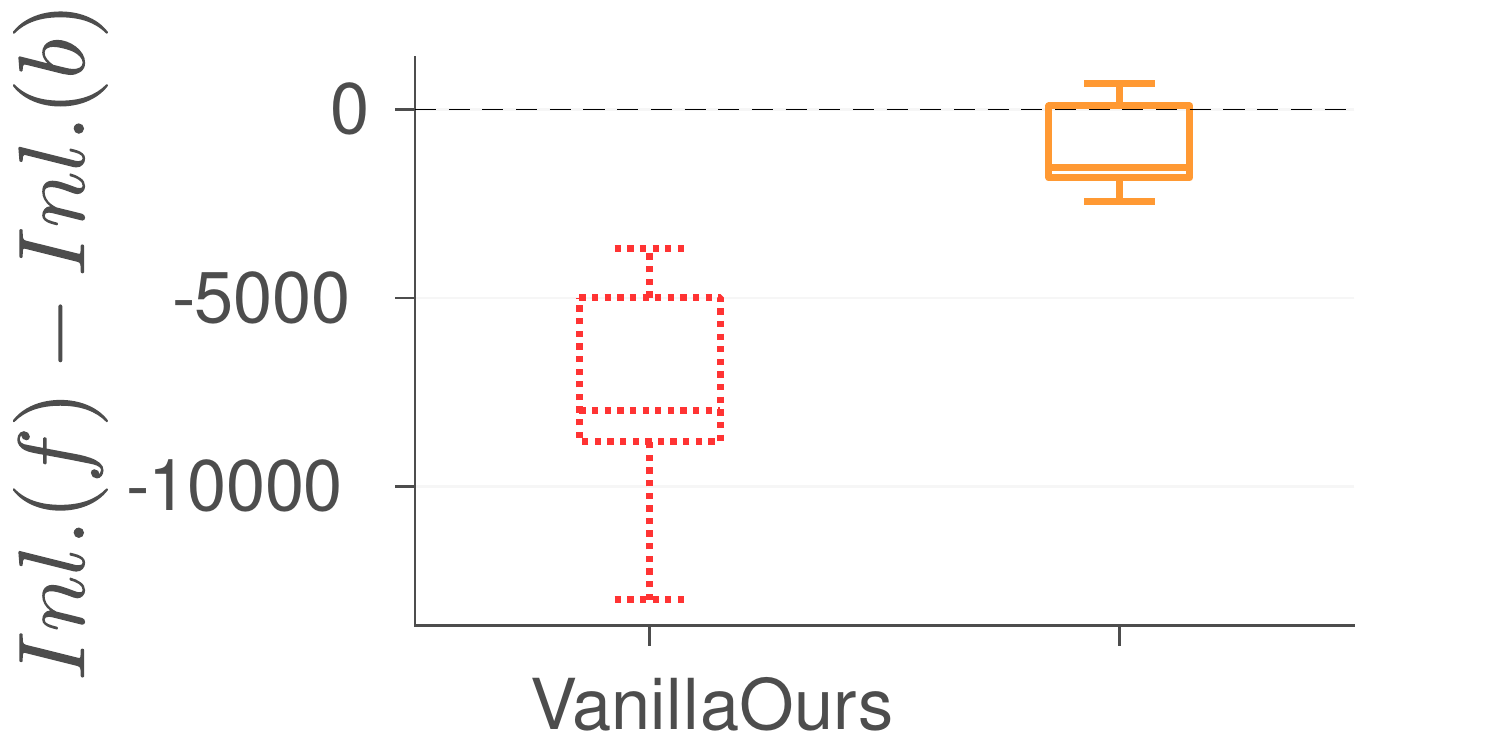}}
  \hfill
  \caption{\textbf{Other motion bias metrics}. We show the differences on the total number [\textit{\#}] of map points (\textit{Pt}), local keyframes (\textit{Key}), and observation inliers (\textit{Inl.}) between forward (\textit{f}) and backward (\textit{b}) passes. As before, our contributions effectively minimize the differences across the three indicators, providing evidence of reduced motion bias.}
  \label{fig:graph}
\end{figure}

An alternative approach to evaluate motion bias without relying on camera trajectory ground truth is by comparing key statistics of the SLAM graph. Figure \ref{fig:graph} illustrates how our proposed approaches effectively reduce the differences across three indicators, providing evidence of reduced motion bias. 

\section{Results}
\label{sec:evaluation}
We selected four publicly available datasets, encompassing a wide range of potential sources for motion bias, including \textit{indoor/outdoor} environments, \textit{walking/driving} scenarios, \textit{high/low} video frequencies, and \textit{strong/soft} forward/backward motion. 

Table \ref{tab:datasets} provides an overview of these dataset characteristics and shows the errors and motion bias metrics obtained by our deterministic implementation of ORB-SLAM2. Note first how motion bias appear in the four datasets, evidencing that it is a general problem in SLAM. With our contributions, we successfully eliminate motion bias across all datasets, effectively balancing the accuracy between forward and backward passes. Notably, in the KITTI dataset, motion bias is completely eradicated, leading to a significant improvement in the accuracy of the backward pass. For detailed results per sequence, refer to Table \ref{tab:kitti_sequences}.

\begin{table}[]
\centering
\begin{minipage}{0.85\linewidth}
\resizebox{1.0\columnwidth}{!}{%
\begin{tabular}{lllllcccccc}
                &         &      & & & \multicolumn{2}{c}{\cellcolor[HTML]{E7FEE9} $e_r$ (f)}  &  \multicolumn{2}{c}{\cellcolor[HTML]{DFE5EF}$e_r$ (b)}& \multicolumn{2}{c}{\cellcolor[HTML]{FEF4D2}$e_r$ (f) - $e_r$ (b)}\\ 
\cline{4-11}
\textbf{Dataset}                  &         &      & \textbf{Hz}  & \textbf{Fwd/Bwd} & Van. & Ours & Van. & Ours & Van. & Ours\\ \hline
%\rowcolor[HTML]{ECF4FF} 
TUM Mono VO \cite{engel2016photometrically}   & Indoor  & Walk & 40  & Medium  & \textbf{6.10} & 6.74 & \textbf{1.89} & 6.86 & 4.51 & \textbf{3.93}\\
RGB-D SLAM \cite{sturm12iros}                 & Indoor  & Walk & 30  & Soft    & 0.07 & \textbf{0.06} &  \textbf{0.03} & 0.05 &0.02 & \textbf{0.01}\\
%\rowcolor[HTML]{ECF4FF} 
EuRoC \cite{burri2016euroc}     & Indoor  & Drone & 20  & Soft    & \textbf{0.07} & 0.16 &0.09& \textbf{0.07}&0.04&\textbf{0.01}\\
KITTI \cite{geiger2013vision}                        & Outdoor & Car  & 10  & Strong  & 76.42  & \textbf{72.72} & 132.24 & \textbf{71.40}  & 65.67 &  \textbf{6.94}
\end{tabular}
}
\end{minipage}
\caption{\textbf{Final Evaluation}. Our approach eliminates motion bias in the four datasets (third column in orange) while keeping global accuracy constant. As expected the reduction of motion bias is stronger in Kitti and TUM Mono VO datasets where the motion occurs mainly in the forward/backward direction of the camera. Values
are the medians after outlier removal.}
\label{tab:datasets}
\end{table}

\begin{table}[]
\centering
\begin{minipage}{0.75\linewidth}
\resizebox{1.0\columnwidth}{!}{%
\begin{tabular}{ccccccccccc}
Sequence       & 00 & 02 & 03 & 04 & 05 & 06 & 07 & 08 & 09 & 10 \\ \hline
\cellcolor[HTML]{E7FEE9} $e_r (f)$ Van.  & \textbf{128.3} & 38.0 & 9.8 &  \textbf{4.1} &  \textbf{65.1} &   \textbf{93.2} &   \textbf{40.1} &  123.1 &  100.4 &   23.4  \\
\cellcolor[HTML]{DFE5EF}$e_r (b)$ Van.  &  254.9 &  179.3 & 6.9 & 0.9 & 115.3 & 111.7 & 53.3 & 183.2 & 120.3 & 29.9\\
\cellcolor[HTML]{FEF4D2}$e_r (f) - e_r (b)$ Van. & -126.6 & -141.3 &    2.9 &    \textbf{3.2} &  -50.1 & -18.4 &  -13.1 &  -60.0 &  -19.9 &   -6.5  \\\hline
\cellcolor[HTML]{E7FEE9}$e_r (f)$ Ours &   140.1 &    \textbf{27.1 }&    \textbf{4.1} &    4.8 &   72.0 &   96.0 &   43.6 &  \textbf{102.6}&  \textbf{71.6} &  \textbf{21.0}    \\
\cellcolor[HTML]{DFE5EF}$e_r (b)$ Ours  & \textbf{136.6} &  \textbf{41.8} &   \textbf{2.0} &   \textbf{0.7} &  \textbf{66.2} & \textbf{100.7} &  \textbf{38.9} & \textbf{102.9} &  \textbf{59.4} &  \textbf{19.3}    \\
\cellcolor[HTML]{FEF4D2}$e_r (f) - e_r (b)$ Ours& \textbf{3.5} &  \textbf{-14.7} &    \textbf{2.0} &    {4.1} &    \textbf{5.7} &   \textbf{-4.6} &    \textbf{4.7} &   \textbf{-0.2} &   \textbf{12.2} &  \textbf{1.68}
  
\end{tabular}
}
\vspace{0.2cm}
\end{minipage}
\caption{\textbf{Per-sequence results in KITTI}. Our motion-bias-free implementation removes the bias ($e_r (f) - e_r (b)$) between forward and backward passes in most sequences.}
\label{tab:kitti_sequences}
\end{table} 

\vspace{-0.44cm}

\section{Conclusion \& Future Work}
\label{sec:conclusion}
In this paper, we are motivated by the significant impact of \textbf{motion bias} on state-of-the-art feature-based SLAM systems. Our findings reveal substantial inconsistencies in performance, not only between forward and reverse directions of travel but also in terms of which direction exhibits better performance.

Our study reveals that the primary factor contributing to the observed inconsistencies is the \textbf{outlier rejection} process. To address this issue, we propose an alternative policy and introduce a \textbf{new model for visual covariances} that effectively removes motion bias. By implementing this and other contributions in ORB-SLAM2, we demonstrate a substantial enhancement in the consistency between forward and backward motion, along with improved average trajectory error in both directions, across four diverse datasets. This novel approach yields more trustable and generalizable performance, which is of great significance in critical applications where consistency of performance is crucial for usability and practicality.

Future work should focus on investigating how outlier management techniques can effectively eliminate motion bias while maintaining the highest level of trajectory accuracy (Section \ref{sec:ablation}). Additionally, we will further explore the impact of image resolution (Section \ref{sec:relatedWork}) and keyframe management strategies (Section \ref{sec:implementationDetails}) on inducing motion bias.

\bibliography{fontanbib, jcivera}
\end{document}